%% file: neurips_2019.tex
\documentclass{article}

     \PassOptionsToPackage{square, comma, sort&compress, numbers}{natbib}

     \usepackage[preprint]{neurips_2019}

\usepackage[utf8]{inputenc} 
\usepackage[T1]{fontenc}    
\usepackage{hyperref}       
\usepackage{url}            
\usepackage{booktabs}       
\usepackage{amsfonts}       
\usepackage{nicefrac}       
\usepackage{microtype}      
\usepackage{graphicx}
\usepackage{float}
\usepackage{url}
\usepackage{amsmath}
\usepackage{comment}
\usepackage{algorithm}  
\usepackage{algorithmic} 
\usepackage{diagbox}
\input{math_commands.tex}

\usepackage{color}
\title{Beneficial perturbation network for continual learning}

\author{%
  Shixian Wen \\
  Department of Computer Science\\
  University of Southern California\\
  Los Angeles, CA 90089\\
  \texttt{shixianw@usc.edu} \\
  \And
  Laurent Itti \\
  Department of Computer Science \\
  University of Southern California\\
  Los Angeles, CA 90089\\
  \texttt{itti@usc.edu} \\
}

\begin{document}

\maketitle

\begin{abstract}
 Sequential learning of multiple tasks in artificial neural networks using gradient descent leads to catastrophic forgetting, whereby previously learned knowledge is erased during learning of new, disjoint knowledge. Here, we propose a fundamentally new type of method - Beneficial Perturbation Network (BPN). We add task-dependent memory (biasing) units to allow the network to operate in different regimes for different tasks. We compute the most beneficial directions to train these units, in a manner inspired by recent work on adversarial examples. At test time, beneficial perturbations for a given task bias the network toward that task to overcome catastrophic forgetting. BPN is not only more parameter-efficient than network expansion methods, but also does not need to store any data from previous tasks, in contrast with episodic memory methods. Experiments on variants of the MNIST, CIFAR-10, CIFAR-100 datasets demonstrate strong performance of BPN when compared to the state-of-the-art.  
\end{abstract}

\section{Introduction}
Continual learning is central to designing general A.I. systems that can learn new tasks sequentially without forgetting previous tasks. However, current deep learning models based on stochastic gradient descent severely suffer from catastrophic forgetting \citep{french1999catastrophic,mccloskey1989catastrophic}, in that they often forget all previous tasks after training each new one. Two major types of methods have been proposed to alleviate this catastrophic forgetting. 

\begin{figure}[]
	\begin{center}
		\includegraphics[height = 8cm]{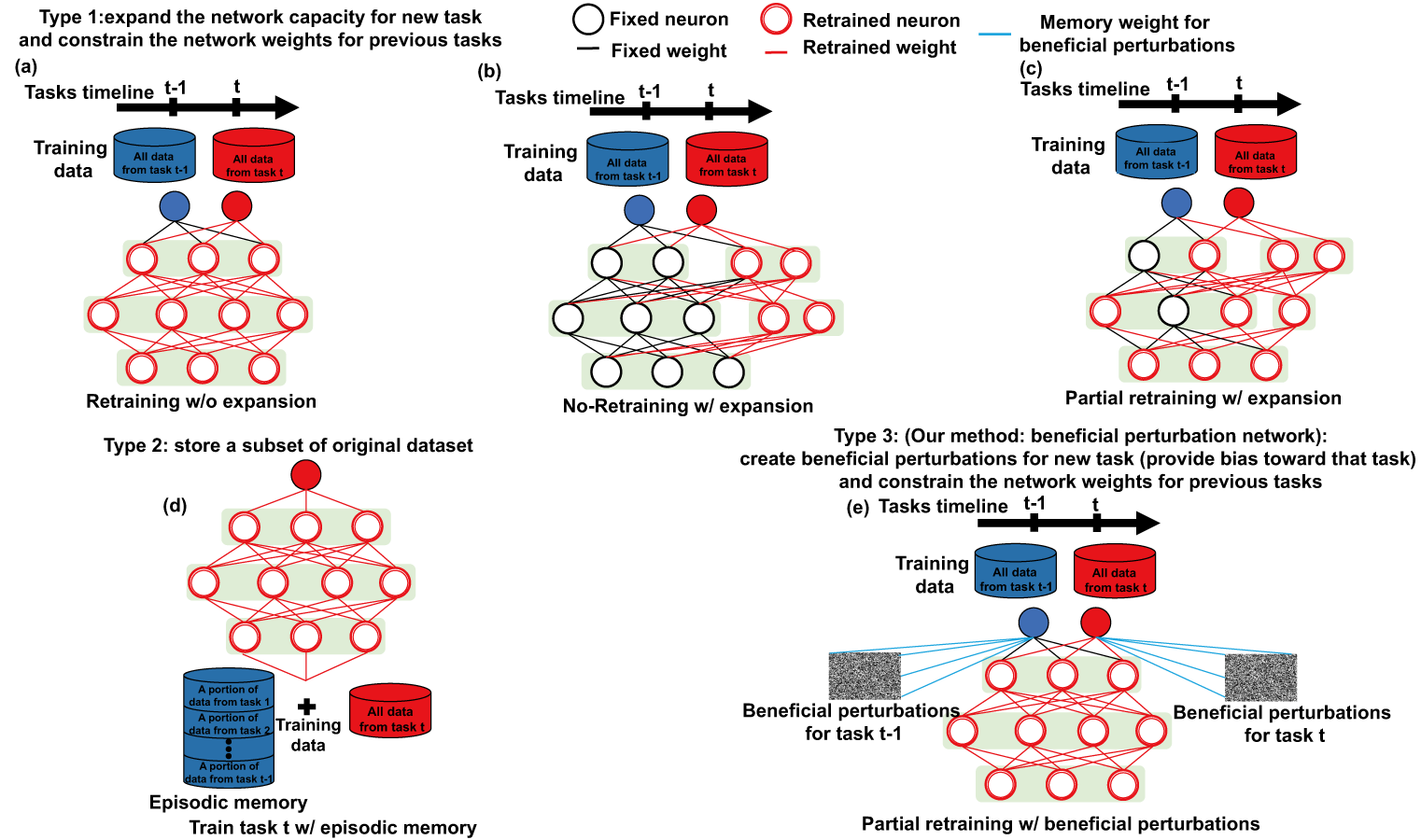}
		\caption{{\bf Concept:} Type 1 expanding and retraining methods (a-c): (a) Retraining models such as elastic weight consolidation retrains the entire network learned on previous tasks while using a regularizer to prevent drastic changes from the original model. (b) Expanding models such as progressive neural network expands the network for new task t without any modifications of the network weights for previous tasks. (c) Expanding model with partial retraining such as dynamically expandable networks expands the network for new task t with partial retraining on the network weights for previous tasks. Type 2 episodic memory methods (d): Methods such as Gradient Episodic Memory stores a subset of the original dataset from previous tasks into the episodic memory and replays them with new data on the training of new tasks. Type 3 beneficial perturbation methods (e):  Beneficial perturbation network creates beneficial perturbations which provide bias toward that task and retrains the normal weights learned from previous tasks using elastic weight consolidation.}
		\label{fig:concept}
	\end{center}
\end{figure}

{\bf Type 1: dynamic network expansion for the new task with constrained network weights for previous tasks.} When new tasks are highly relevant to the old ones, using a regularizer that prevents drastic parameter changes is sufficient to find a good solution for the new tasks (Fig.~\ref{fig:concept}a). Methods such as learning without forgetting \cite{li2017learning}, elastic weight consolidation (EWC) \cite{kirkpatrick2017overcoming}, and incremental moment matching \cite{lee2017overcoming} have deployed this strategy. However, if the new tasks are largely different from the old ones, e.g., when previous tasks are to classify images of fruits and the new task is to classify images of automobiles, the learned features cannot accurately represent the new tasks, and additional neurons need to be introduced to the network. Additional neurons account for the new features that are necessary for the new task. Progressive neural network \citep{rusu2016progressive} retains a pool of pretrained networks throughout training of previous tasks, and expands the network alongside the pretrained networks to learn a new task (Fig.~\ref{fig:concept}b). Dynamically expandable networks (Fig.~\ref{fig:concept}c) \citep{yoon2018lifelong} combine the above two methods by selectively retraining the old networks and expanding their capacity when necessary. The tradeoff is that the expansion methods increase the network parameters linearly with the number of tasks.

{\bf Type 2: using an episodic memory to store a subset of the original dataset from previous tasks and replaying them to maintain predictions invariant during the training of a new task.} For this type of method (Fig.~\ref{fig:concept}d), the goal is to mimic the distribution of data from previous tasks by storing a fraction of the original dataset into the episodic memory. For each new task, the network avoids catastrophic forgetting by learning from both new data and replay of old data.  Several variations have been proposed such as iCARL and Gradient Episodic Memory (GEM) \citep{rebuffi2017icarl,lopez2017gradient}. Instead of storing individual data into the episodic memory, one can train generative models to reproduce the lost data and labels. Several variations have been proposed, leveraging autoencoders \citep{rannen2017encoder} or generative adversarial networks \citep{rios2018closed}. However, the replay of old data increases training time for new tasks. It is neither memory efficient nor parameter efficient because one has to store the old data or has to build a generative model for each new task.

Here, we propose a fundamentally new type of method - {\bf Beneficial Perturbation Network (BPN)} (Fig.~\ref{fig:concept}e). To understand beneficial perturbations, we first revisit the meaning of adversarial perturbations in adversarial examples. With adversarial examples \citep{szegedy2013intriguing}, it has been shown that adversarial perturbations (calculated from a specific class) added to input images can bias a network to misclassify the perturbated input images into that specific class. Here, we leverage this idea, but, instead of adding input "noise" (adversarial perturbations) calculated from other classes to force the network into misclassification, we can add "noise" (beneficial perturbations) calculated from the input's own correct class to assist correct classification. Instead of adding perturbations to the input images, we introduce additional neural network units (memory units) to store the beneficial perturbations for each task, which provide bias towards that task.  Memory units and their weights (memory weights) are trained independently for each task, in addition to continued updating of the normal weights of the network (weights for normal neurons, constrained by elastic weight consolidation). At test time, memory units for a given task are activated and they bias the network for that task, thereby overcoming any corruption of the normal weights which are shared by all tasks. As further explained below, and as a result of our inspiration from adversarial attacks, we find that training the memory units using the Fast Gradient Sign Method (FGSD) yields the highest performance \citep{tramer2017space}. We further show that training the memory units using gradient descent fails, thus the performance of our approach is not simply due to adding more dimensions.

We validate our beneficial perturbation network for lifelong learning on multiple public datasets (incremental MNIST, incremental CIFAR-10 and incremental CIFAR-100), on which it achieves comparable or better performance than the state-of-the-art. For each task, by adding beneficial perturbations to the last layer of a 5-layer fully connected network, we only introduce 10.0\% parameters increase, compared to 100\% parameters increase for models that train a separate network for each task, and 11.9\% - 60.3\% for the dynamically expandable network. In addition, our model does not need a large episodic memory to store data from the previous tasks and to replay them during the training of new tasks. Thus, the beneficial perturbation network has a promising future to alleviate catastrophic forgetting compared to the other two types of methods. 

\section{Background and Motivation}
 {\bf Biological inspiration: memory formation and retrieval in Hippocampus (HPC):}
In the human brain, the hippocampus \citep{bakker2008pattern} encodes detailed information in Cornu Ammonis 3 (CA3), which does pattern separation, and transforms this information into abstract high-level information, then relayed to Cornu Ammonis 1 (CA1), which does pattern completion. During weight consolidation \citep{lesburgueres2011early,squire1995retrograde,frankland2005organization}, the HPC fuses different features from different tasks into a coherent memory trace. Over days to weeks, as memories mature, HPC progressively stores permanent abstract high-level long-term memories to remote memory storage (neocortical areas). HPC can maintain and mediate their retrieval independently when the specific memory is in need.   

 {\bf Adversarial perturbations, directions vs beneficial perturbations, directions}
 
 By adding a carefully computed “noise” (adversarial perturbations) to an input picture, without changing the neural network, one can force the network into misclassification.
The noise is usually computed by backpropagating the gradient in a so-called “\textbf{\em adversarial direction}” such as by using the fast gradient sign method (FGSD) \citep{tramer2017space}.

\begin{figure}[H]
	\begin{center}
		\includegraphics[height = 3cm]{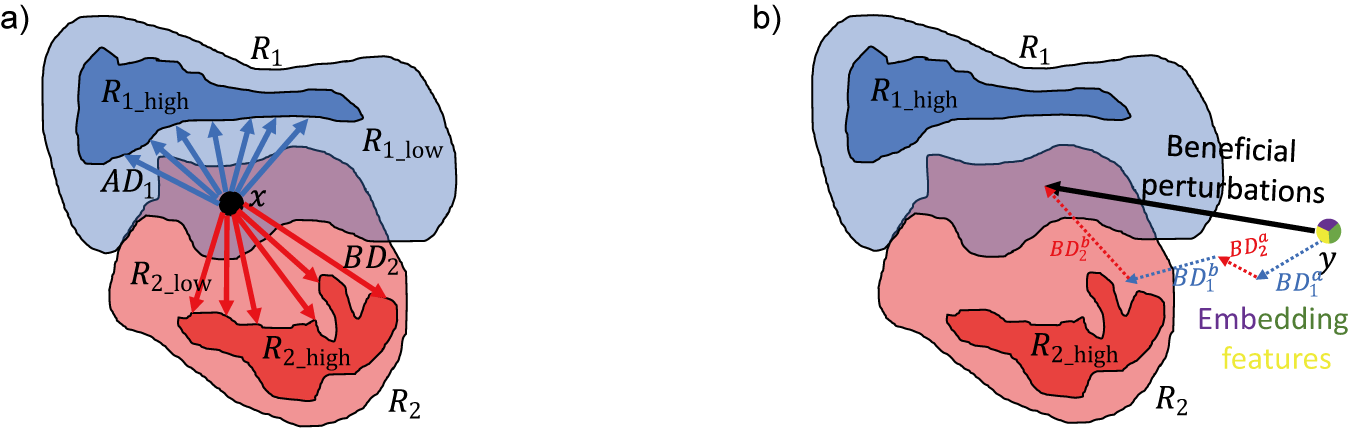}
		\caption{ (a) {\bf Adversarial direction (AD) vs beneficial direction (BD)}. $R_1$ ($R_2$) is the classification region (region of constant estimated label) of digit 1 (digit 2) from the MNIST dataset. Subregion $R_{1\_high}$ ($R_{1\_low}$) is the high (low) classification region of digit 1, and likewise for $R_{2\_high}$ ($R_{2\_low}$) for digit 2. The data point $x$ is a clear input image of digit 2 that lies in the intersection of $R_{1\_low}$ and $R_{2\_low}$. $AD_1$ is an adversarial direction for class digit 1. $BD_2$ is a beneficial direction for class digit 2. $x+AD_1$ crosses the decision boundary of $R_2$ and towards $R_{1\_high}$. $x+BD_2$ is moved towards $R_{2\_high}$. (b) {\bf Beneficial perturbations for the task that has two input classes (digit 1, 2).} $BD_1$ ($BD_2$) is a beneficial direction for class digit 1 (2). The network has been trained on two images from digit 1 ($1^a$ and $1^b$) and two images from digit 2 ($2^a$ and $2^b$). Thus, the beneficial perturbations for this task are the vector ($BD_1^{a} + BD_2^{a} + BD_1^{b} + BD_2^{b}$). Point $y$ is the activations of normal neurons of each layer in the neural network that has been corrupted by other tasks. The contaminated point y alone cannot be correctly classified as digits 1 or 2 because it lies outside of the classification region ($R_1$ or $R_2$). In test time, adding beneficial perturbations to the embedding features move toward the classification regions (intersection of $R_1$ and $R_2$) that have non-negligible network response for all classes of this task. Thus, beneficial perturbations bias the network toward this task.}
		\label{fig:beneficial_perturbations}
	\end{center}
\end{figure}

Attack and defense researchers usually view adversarial examples as a curse of neural networks, but we view it as a gift to solve catastrophic forgetting. In Fig.~\ref{fig:beneficial_perturbations}a, the adversarial perturbations are obtained by backpropagating from the class digit 1 to the input space, following any of the available adversarial directions (blue arrows $AD_1$). Adding adversarial perturbations to the input image can be viewed as adding an adversarial direction vector (toward the high confidence classification region of digit 1) to the clear (non-perturbated) input image of digit 2 ($x$). The resulting vector crosses the decision boundary of the classification region of digit 2 ($R_2$). Thus, adversarial perturbations can force the neural network into misclassification, here from digit 2 to digit 1. Because the dimensionality of adversarial directions is around 25 for MNIST \citep{tramer2017space}, when we project them into a 2D space, we use the fan-shaped blue arrows to depict those dimensions. 

Instead of adding input "noise" (adversarial perturbations) calculated from other classes to force the network into misclassification, we can add "noise" (beneficial perturbations) calculated by the input's own correct class to assist correct classification. In Fig.~\ref{fig:beneficial_perturbations}a, the beneficial perturbations are obtained by backpropagating from the target class (digit 2) to the input space with the FGSD method. The beneficial perturbations are vectors that point toward the direction of high confidence classification region of digit 2 (red arrows). Adding beneficial perturbations to the data point $x$ (here, with true label digit 2) allows the beneficial perturbed image to enter the high confidence classification region of digit 2. Thus, it forces the network to classify the beneficial perturbed image as class digit 2 with high confidence. To overcome catastrophic forgetting, we create some beneficial perturbations for each task, stored in task-dependent memory units. They can bias the network toward that particular task, even though the shared knowledge (normal weights) becomes contaminated by other tasks (Fig.~\ref{fig:beneficial_perturbations}b, Fig.~\ref{fig:explanation_structure}). The beneficial perturbations for each task are created by aggregating  the beneficial direction vectors sequentially for each class through mini-batch backpropagation. For example, during the training of task 1, the network has been trained on two images from digit 1 ($1^a$ and $1^b$) and two images from digit 2 ($2^a$ and $2^b$). The beneficial perturbations for task 1 are the summation of the beneficial directions calculated from each image ($BD_1^{a} + BD_2^{a} + BD_1^{b} + BD_2^{b}$ in Fig.~\ref{fig:beneficial_perturbations}b). After adding beneficial perturbations to the embedding features (point y: activations of normal neurons of each layer in the neural network, which might be corrupted by other tasks), beneficial perturbations help the embedding features move toward the classification regions that have non-negligible network response for all classes of this task (the intersection of $R_1$ and $R_2$ in Fig.~\ref{fig:beneficial_perturbations}b). Thus, beneficial perturbations bias the neural network toward that task (Fig.~\ref{fig:explanation_structure}). In summary, we create "biased" knowledge (beneficial perturbations) for each task, stored in task-dependent memory units, that can bias the network toward that particular task, even though the common knowledge (normal weights) becomes corrupted by other tasks. We propose that this "biased" knowledge (beneficial perturbations for each task), is a representation of \textbf{\em long-term memory} in neural networks, although, we do not know how our brain represents permanent long-term memories. In test time, the beneficial perturbation network operates similar to the hippocampus of our human brain. It activates the corresponding long-term memory and biases the network toward that task.

\section{Beneficial Perturbation Network (BPN)}

The beneficial perturbation network has task-dependent memory units (\(\mM_{task}\in R^{K{\times}H}\), K is the number of memory units, H is the hidden size of a memory unit) in the last layer to store the beneficial perturbations. The stored beneficial perturbations can bias the neural network outputs to each task after training on different sequential tasks. We use a scenario with two tasks to illustrate the method.

\begin{figure}[]
	\begin{center}
		\includegraphics[width=1\linewidth]{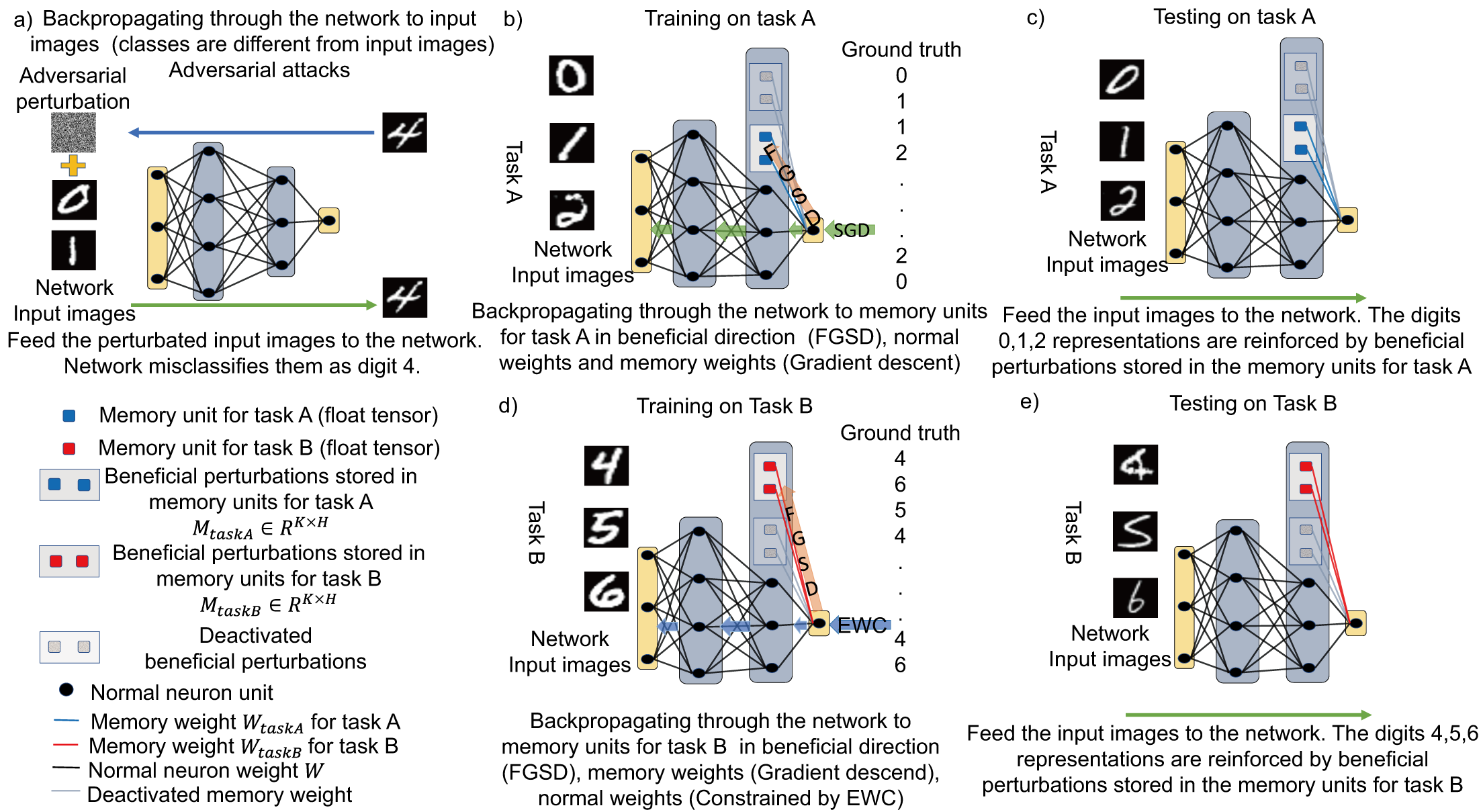}
		\caption{ (a) classical adversarial examples. (b-e): Beneficial perturbation network with 2 tasks.  (b) Train on task A. (c) Test on task A. (d) Train on task B. (e) Test on task B.}
		\label{fig:explanation_structure}
	\end{center}
\end{figure}

First, during training of task A (Fig.~\ref{fig:explanation_structure} b), we consider MNIST input images 0,1,2. We use the forward and backward rules (Alg.~\ref{alg:FORTA}, Alg.~\ref{alg:BACTA} shown in blue) to update the BPN. We update the memory units ($\mM_{task_A}$) in the beneficial direction (FGSD) as $\epsilon sign(\nabla_{\mM_{task_A}} L(\mM_{task_A},\vy_{target}))$, where $\mM_{task_A}$ are memory units for task A,  $\vy_{target}$ are the current inputs' true classes (MNIST input images 0,1,2). This allows the memory units for task A ($\mM_{task_A}$) to store the beneficial perturbations for task A (a vector towards the regions that have non-negligible network response for MNIST digits 0,1,2). Deactivating memory units that are not for the current task (Task B), and activating memory units for the current task (Task A), ensures that the beneficial perturbations are not being corrupted by other tasks (Task B) and can only be modified by the current training task (Task A). 

After training of task A, during training of task B (Fig.~\ref{fig:explanation_structure} d), we consider MNIST input images 4,5,6. We again use the forward and backward rules (Alg.~\ref{alg:FORTA}, Alg.~\ref{alg:BACTA} shown in red) to update the BPN. In the loss function of Alg.~\ref{alg:BACTA}, $\lambda \mF_i(\mW_i-\mW^{A*}_{i})^2)$ is the EWC constraint we apply to the normal weights to maintain a joint probability distribution between task A and task B, where $i$ labels each parameter, $\mF_i$ is the Fisher information matrix, $\lambda$ sets how important the old task is compared to the new one, $\mW_i$ are normal weights, and $\mW_{i}^{A*}$ are optimal normal weights after training on task A. We update our memory units ($\mM_{task_B}$) in the beneficial direction (FGSD) as $\epsilon sign(\nabla_{\mM_{task_B}} L(\mM_{task_B},\vy_{target}))$, where $\mM_{task_B}$ are memory units for task B,  $\vy_{target}$ are the current inputs' true classes (MNIST input images 4,5,6). This allows our memory units for task B ($\mM_{task_B}$) to store the beneficial perturbations for task B (a vector towards the regions that have non-negligible network response for MNIST digits 4,5,6).  

After training of task B, we test accuracy for task A on a test set. The forward rules are in Alg.~\ref{alg:FORTA} shown in blue (Fig.~\ref{fig:explanation_structure} c). During testing of task A, although the common knowledge (normal weights $\mW$) has been contaminated by task B, the integrity of memory units for task A that store the beneficial perturbations still can bias the network outputs to task A. During testing of task B, the forward rules are Alg.~\ref{alg:FORTA} shown in red (Fig.~\ref{fig:explanation_structure} e). The memory units for task B can bias the network outputs to task B, in case the common knowledge (normal weights) is further modified by later tasks. In scenarios with more than two tasks, the forward and backward algorithms for later tasks are the same as for task B, except that they will update their own memory units, and turn off memory units for other tasks.

\begin{algorithm}[tb]
\caption{forward rules for task {\color{blue}A} ({\color{red}B}) }
\label{alg:FORTA}

\begin{algorithmic}

\STATE {\underline {For each layer that has memory units do:}}  
\STATE {\quad Activate memory units (task {\color{blue}A} ({\color{red}B}): $\mM_{task_{{\color{blue}A}({\color{red}B})}}$) for the current task }
\STATE{\quad Deactivate memory units (task {\color{blue}B} ({\color{red}A}): $\mM_{task_{{\color{blue}B}({\color{red}A})}}$) not for the current task }
\STATE{\quad{\bfseries Input:}\hspace{0.15cm} $\mM_{task_{{\color{blue}A}({\color{red}B})}}$ \textemdash
\hspace{0.08cm} memory units for task {\color{blue}A} ({\color{red}B})}
\STATE{\quad\quad\quad\quad\hspace{0.15cm}$\mX_{activations}$\textemdash
\hspace{0.08cm} activations from the last layer}
\STATE{\quad{\bfseries Output:} $\mY = \mW_{task_{{\color{blue}A} ({\color{red}B})}}  \mM_{task_{{\color{blue}A}({\color{red}B})}} + \mW  \mX_{activations} + \mathbf{bias}$ \hfill {\color{green}// activations for the next layer}}
\STATE{\quad\quad\qquad \hspace{0.13cm} where: $\mW_{task_{{\color{blue}A}({\color{red}B})}}$\textemdash
\hspace{0.08cm} memory weights for task {\color{blue}A}({\color{red}B}). \hfill $\mW$\textemdash
\hspace{0.08cm} normal neuron weights.}

\STATE{}

\STATE {\underline {For each layer that does not have memory units do (fully connected layer):}}
\STATE{\quad{\bfseries Input:} \hspace{0.13cm} $\mX_{activations}$\textemdash
\hspace{0.08cm} activations from last layer}
\STATE{\quad{\bfseries Output:} $\mY =\mW  \mX_{activations} + \mathbf{bias}$ \hfill {\color{green}// activations for the next layer}}
\STATE{\quad\quad\quad\quad\qquad \hspace{0.13cm}where: $\mW$\textemdash
\hspace{0.08cm} normal neuron weights.}
\end{algorithmic}
\end{algorithm}

\begin{algorithm}[tb]
\caption{backward rules for {\color{blue}A} ({\color{red}B})}
\label{alg:BACTA}

\begin{algorithmic}

{\color{blue}
\STATE {\underline {For Task A:}}
\STATE{\quad minimizing loss function: $L(\mX_{dataA},\mM_{task_A},\mW,\mW_{task_A})$}
\STATE{\quad \quad where: $\mX_{dataA}$\textemdash
\hspace{0.08cm} data for task A.\quad $\mM_{task_A}$  \textemdash
\hspace{0.08cm} memory units for task A.}
\STATE{\quad\quad\quad\quad\quad $\mW$\textemdash
\hspace{0.08cm} normal neuron weights.\quad $\mW_{task_A}$  \textemdash
\hspace{0.08cm} memory weights for task A.}}
\STATE{}
{\color{red}
\STATE {\underline {For Task B:}}
\STATE{\quad minimizing loss function: $L(\mX_{dataB},\mM_{task_B},\mW,\mW_{task_B})+ \sum_{i} \lambda \mF_i(\mW_i-\mW^{A*}_{i})^2$}
\STATE{\quad \quad where: $\mX_{dataB}$\textemdash
\hspace{0.08cm} data for task B.\quad $\mM_{task_B}$  \textemdash
\hspace{0.08cm} memory units for task B.}
\STATE{\quad\quad\quad\quad\quad $\mW$\textemdash
\hspace{0.08cm} normal neuron weights.\quad $\mW_{task_B}$  \textemdash
\hspace{0.08cm} memory weights for task B.}
\STATE{\quad\quad\quad\quad\quad $i$\textemdash
\hspace{0.08cm} labels each parameter.\quad $\mF_{i}$  \textemdash
\hspace{0.08cm} Fisher information matrix for parameter i.}
\STATE{\quad\quad\quad\quad\quad $W_i$\textemdash
\hspace{0.08cm} normal weight i.\hfill $\mW^{A*}_{i}$  \textemdash
\hspace{0.08cm} optimal normal weight i after training on task A.}
}
\STATE{}
\STATE {\underline {For each layer that has memory units do:} } 
\STATE {\quad Activate memory units (task {\color{blue}A} ({\color{red}B}): $\mM_{task_{{\color{blue}A} ({\color{red}B})}}$) for the current task }
\STATE{\quad Deactivate memory units (task {\color{blue}B} ({\color{red}A}): $\mM_{task_{{\color{blue}B}  ({\color{red}A})}}$) not for the current task }
\STATE{\quad{\bfseries Input:}\hspace{0.15cm} $\mathbf{Grad}$ \textemdash
\hspace{0.08cm} Gradients from the next layer}

\STATE{\quad {\bfseries output:} $\mathbf{dW_{task_{{\color{blue}A}({\color{red}B})}}} = \mathbf{Grad}.dot(\mM^T_{task_{{\color{blue}A}({\color{red}B})}})$  \hfill {\color{green}// gradients for memory weights of task {\color{blue}A} ({\color{red}B})}}
\STATE{ \quad  $\hspace{33pt}$   $\mathbf{dM_{task_{{\color{blue}A}({\color{red}B})}}} = \epsilon\; sign\;(\mathbf{W^T_{task_{{\color{blue}A}({\color{red}B})}}}.dot(\mathbf{Grad}))$ }
\STATE{\hfill  {\color{green}// gradients for memory units of task {\color{blue}A} ({\color{red}B}) using FGSD method}}
\STATE{ \quad $\hspace{33pt}$   $\mathbf{dW} = \mathbf{Grad}.dot(\mathbf{X^T_{activations}})$\hfill {\color{green}// gradients for normal weights}}
\STATE{ \quad  $\hspace{33pt}$   $\mathbf{dX_{activations}} = \mathbf{W^T}.dot(\mathbf{Grad})$ \hfill {\color{green}// gradients for activations to last layer}}
\STATE{ \quad  $\hspace{33pt}$   $\mathbf{dbias} = \mathbf{Grad}.sum(0).squeeze(0)$ \hfill {\color{green}// gradients for bias}}
\STATE{}
\STATE {\underline {For each layer that does not have memory units do (fully connected layer):}}  
\STATE{\quad{\bfseries Input:}\hspace{0.20cm} $\mathbf{Grad}$ \textemdash
\hspace{0.08cm} Gradients from the next layer}
\STATE{ \quad {\bfseries Output:}   $\mathbf{dW} = \mathbf{Grad}.dot(\mathbf{X^T_{activations}})$\hfill {\color{green} // gradients for normal weights}}
\STATE{ \quad  $\hspace{35pt}$   $\mathbf{dX_{activations}} = \mathbf{W^T}.dot(\mathbf{Grad})$ \hfill {\color{green}// gradients for activations to last layer}}
\STATE{ \quad  $\hspace{35pt}$   $\mathbf{dbias} = \mathbf{Grad}.sum(0).squeeze(0)$ \hfill {\color{green}// gradients for bias}}

\end{algorithmic}
\end{algorithm}

\section{Experiment}
{\bf Datasets:}
{\bf 1. Incremental MNIST.} A variant of the MNIST dataset \citep{lecun1998gradient} of handwritten digits with 10 classes, where each task introduces a new set of classes. We consider 5 tasks; each new task concerns examples from a disjoint subset of 2 classes. {\bf 2. Incremental CIFAR-10.} A variant of the CIFAR object recognition dataset \citep{krizhevsky2009learning} with 10 classes. We consider 5 tasks; each new task has 2 classes. {\bf 3. Incremental CIFAR-100.} A variant of the CIFAR object recognition dataset \citep{krizhevsky2009learning} with 100 classes. We consider 10 tasks; each new task has 2 classes.

{\bf Our model and baselines:} {\bf 1. BD + EWC (K x H) :} Beneficial Perturbation Networks. K is the number of memory units. H is the hidden dimension for each memory unit. The memory units are updated in the beneficial direction (BD) using FGSD method. The memory weights are updated in the gradient direction (GD). The normal weights are updated with EWC constraints.  

{\bf 2. Single task Learning (STL).} 5-layer fully-connected neural network, trained for each task separately. STL does not suffer from catastrophic forgetting at all.

{\bf 3. Elastic weight consolidation (EWC) \citep{kirkpatrick2017overcoming}.} The loss is regularized to avoid catastrophic forgetting.





{\bf 4. Gradient Episodic Memory with task oracle (GEM (*)) \citep{lopez2017gradient},} GEM uses a task oracle to build a final linear classifier (FLC) per task. The final linear classifier adapts the output distributions to the subset of classes for each task. GEM uses an episodic memory to store a subset of the observed examples from previous tasks. We use notation GEM (*) for the rest of the paper, where * is the size of episodic memory (number of training images stored) for each task.

{\bf 5. GD + EWC (K x H):} The update rules and network structure are the same as BD + EWC (K x H), except the memory units are updated in Gradient direction (GD). This method has the same parameter costs as BD + EWC (K x H). The failure of GD + EWC (K x H) suggests that the good performance of BD + EWC (K x H) is not from the additional dimensions provided by memory units.  

 We trained a fully-connected neural network with 5 hidden layers of 300 ReLU units. We compared our beneficial perturbation network (BD + EWC) to 4 alternatives. For incremental MNIST and incremental CIFAR-10 dataset, we added 200 memory units with 200 hidden dimensions only to the last layer of the 5-layer fully-connected neural network (10.0\% parameter increase). For incremental CIFAR-100 dataset, we added memory units with variable size to the last layer, last two layers, or every layer of the 5 layer fully-connected neural network. The purpose is to show the superior performance of beneficial perturbation network compared GEM under the same memory budgets, especially when a hardware device may require low memory budgets.
\begin{figure}[]
	\begin{center}
		\includegraphics[height = 6cm]{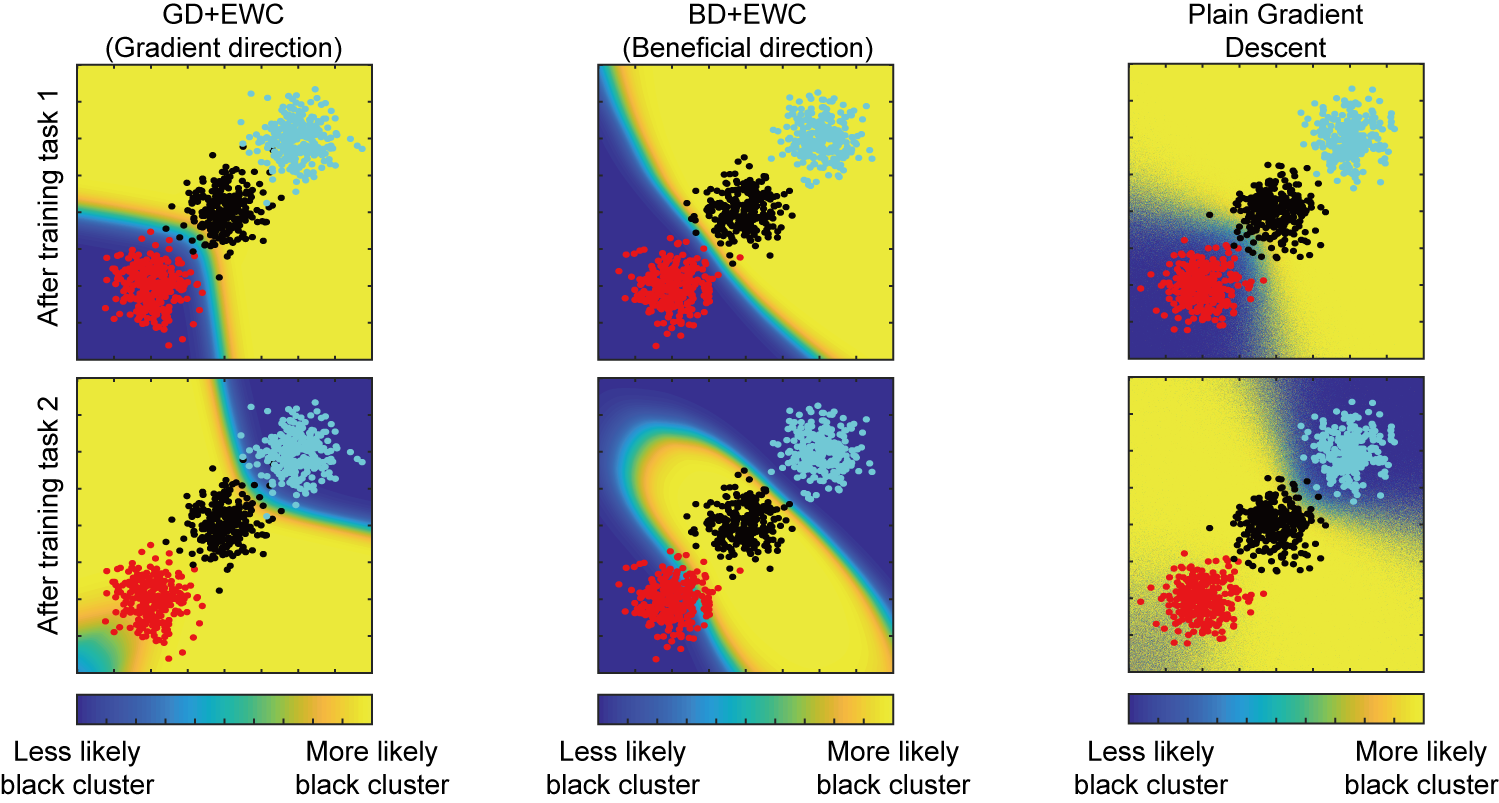}
		\caption{{\bf Visualization of classification regions:} classify 3 randomly generated normal distributed clusters. Task 1: separate black from red clusters. Task 2: separate black from light blue clusters. The yellower (bluer) the heatmap, the higher (lower) the chance the neural network classifies a location as the black cluster. After training tasks 2, only BD + EWC remembered the task 1 by maintaining its decision boundary between the black and red clusters. Both plain gradient descent and GD + EWC forget task 1 entirely.}
		\label{fig:visualization}
	\end{center}
\end{figure}

{\bf Results:}

{\bf The beneficial perturbations can bias the network outputs and maintain the decision boundary.} How to tell whether the advantages of our method are really from the beneficial perturbations and not just from additional dimensions to the neural network? We compare between updating the memory units in the beneficial direction (BD + EWC which comes from beneficial perturbations) and in the gradient direction (GD + EWC, which just comes from the additional dimensions that our memory units and memory weights provide). We use a toy example (classify 3 groups of normal distributed clusters) to demonstrate it and to visualize the decision boundary (Fig.~\ref{fig:visualization}). We randomly generate 3 normal distributed clusters located at different locations. We have two tasks - Task 1: separate the black cluster from the red cluster. Task 2: separate the black cluster from the light blue cluster. The yellower (bluer) the heatmap, the higher (lower) the confidence that the neural network classifies a location into the black cluster. After training the task 2, both plain gradient descent and GD + EWC forget task 1. (dark blue boundary around the red cluster disappeared). However, BD + EWC not only learns how to classify task 2 (clear decision boundary between light blue and black clusters), but also remembers how to classify the old task 1 (clear decision boundary between red and black clusters). Thus, it is the beneficial perturbations which can bias the network outputs and maintain the decision boundary for each task, not just adding more dimensions.

\begin{figure}[]
	\begin{center}
		\includegraphics[height=10.9cm]{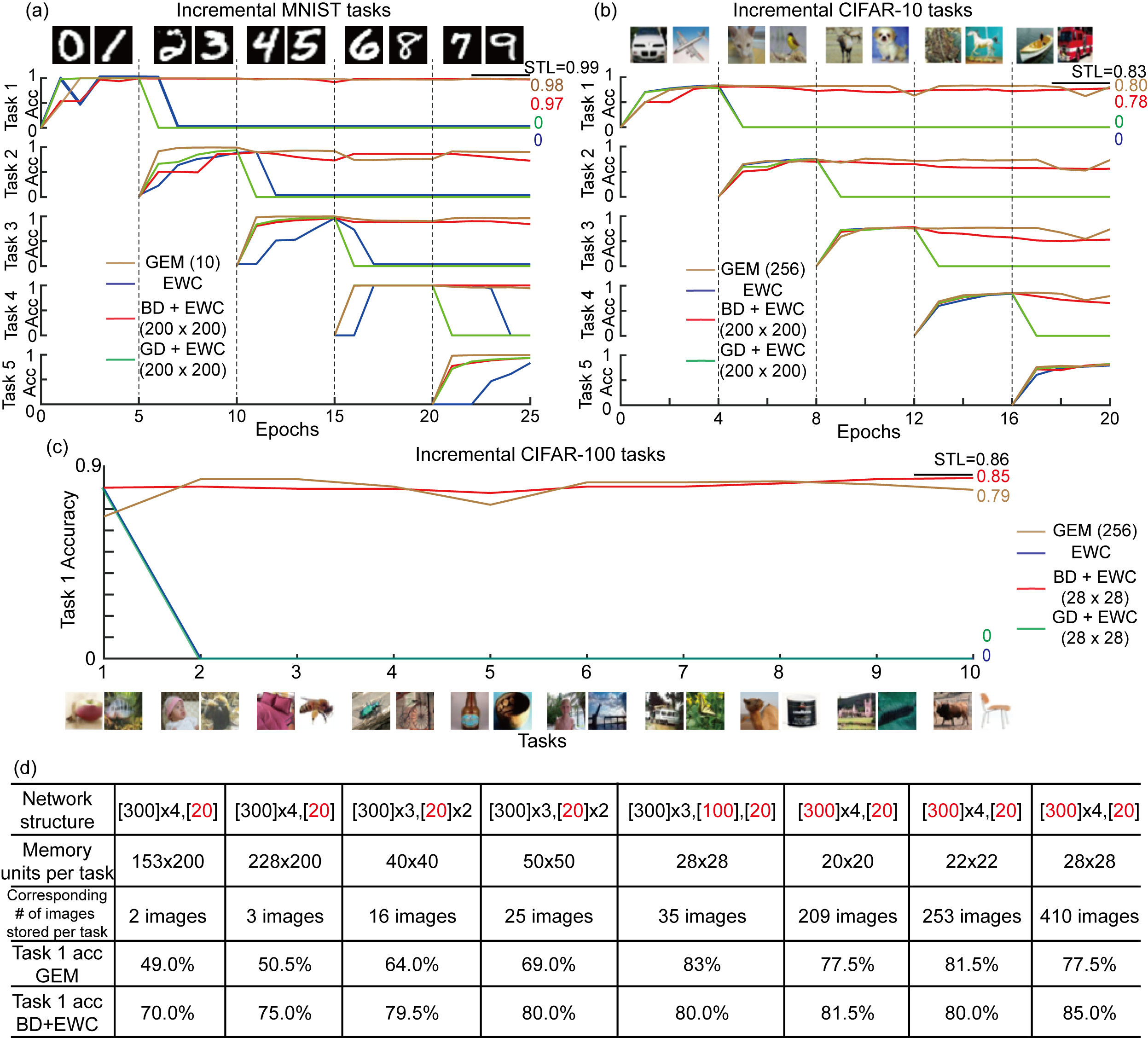}
		\caption{(a) Incremental MNIST tasks. (b) Incremental CIFAR-10 tasks. For a and b, the dashed line indicates the start of a new task. The vertical axis is the accuracy for each task. The horizontal axis is the number of epochs. (c) Incremental CIFAR-100 tasks. The vertical axis is the accuracy for task 1. The horizontal axis is the number of tasks. (d) Task 1 accuracy after sequential training for incremental CIFAR-100 tasks. Each column has the same memory budget for both GEM and BD + EWC methods. In the network structure row, for BD + EWC, the black (red) color represents the fully connected (FC) layer (memory unit layer). For GEM, both red and black color represent the FC layer. Each memory unit layer has the same amount of memory units (memory units per task row). }
		\label{fig:quantative_results}
	\end{center}
\end{figure}

{\bf Quantitative analysis.} Fig.~\ref{fig:quantative_results} summarize the performances for all datasets and methods. STL has the best performance since it trained for each task separately and did not suffer catastrophic forgetting at all. BD + EWC performed slightly worse than STL (2\%,6\%,1\% worse for incremental MNIST, CIFAR-10, CIFAR-100 datasets). BD + EWC achieved comparable or better performance than GEM. On incremental CIFAR-100 dataset, we compared BD + EWC and GEM with same memory budgets (Fig.~\ref{fig:quantative_results}d). When the memory budgets are low (high), BD + EWC outperforms GEM by 32\% (8.82\%). Both methods need a task oracle. GEM uses it to build a final linear classifier to adapt the output distributions to the subset of classes for each task. BD + EWC uses it to activate corresponding memory units. However, the task oracle constrains for BD + EWC is looser than for GEM, since it does not need the final linear classifier for each task. The perturbations stored in the memory units are enough to bias for the task and do this distribution mapping automatically. EWC rapidly decreased to 0\% accuracy. This confirms similar results on EWC performance on incremental datasets \citep{rios2018closed, kemker2017fearnet, parisi2018continual}. GD + EWC has the same additional dimensions as BD + EWC, but GD + EWC failed in the continual learning scenario. The result suggests that it is not the additional dimensions, but the beneficial perturbations which help overcome catastrophic forgetting.

\section{Conclusion}
We proposed a fundamentally new type of method - beneficial perturbation network (BPN) for lifelong learning. Our experiments demonstrate the competitive performance of BPN against the state-of-the-art. Additionally, BPN is more parameter efficient than network expansion methods and does not need a large episodic memory to store any data from previous tasks, compared to episodic memory methods. Through visualization of classification regions and quantitative results, we validated that the beneficial perturbations can bias the network toward the task to overcome catastrophic forgetting. For future research, the organic combinations of 3 types of methods (network expansion method, episodic memory method, beneficial perturbation method) may yield even better performance in more complex scenarios.

\clearpage
\bibliography{neurips_2019}
\bibliographystyle{unsrt}

\end{document}

%% file: math_commands.tex
\usepackage{amsmath,amsfonts,bm}

\def\eqref#1{equation~\ref{#1}}

\def\1{\bm{1}}

\def\vy{{\bm{y}}}

\def\mF{{\bm{F}}}

\def\mM{{\bm{M}}}

\def\mW{{\bm{W}}}
\def\mX{{\bm{X}}}
\def\mY{{\bm{Y}}}

\DeclareMathAlphabet{\mathsfit}{\encodingdefault}{\sfdefault}{m}{sl}
\SetMathAlphabet{\mathsfit}{bold}{\encodingdefault}{\sfdefault}{bx}{n}

